\documentclass[letterpaper, 10 pt, conference]{ieeeconf}  % Comment this line out if you need a4paper

\IEEEoverridecommandlockouts                              % This command is only needed if 
\usepackage{graphics} % for pdf, bitmapped graphics files
\usepackage{epsfig} % for postscript graphics files
\usepackage{mathptmx} % assumes new font selection scheme installed
\usepackage{times} % assumes new font selection scheme installed
\usepackage{amsmath} % assumes amsmath package installed
\usepackage{amssymb}  % assumes amsmath package installed
\usepackage{color}
\usepackage{xcolor}
\usepackage{graphicx}
\usepackage{tabularx}
\graphicspath{ {./images/} }
\usepackage{subcaption}
\usepackage{cite}
\usepackage{mathrsfs}
\usepackage{multirow}
\usepackage{float}
\usepackage{caption}
\usepackage[hidelinks]{hyperref}

\title{\LARGE \bf
ATPPNet: Attention based Temporal Point cloud Prediction Network
}

\author{Kaustab Pal$^{*1}$, 
Aditya Sharma$^{*1}$,
Avinash Sharma$^{2}$,
K. Madhava Krishna$^{1}$% <-this % stops a space
\thanks{$^*$ denotes equal contribution.} 
\thanks{$^1$ are with RRC, IIIT Hyderabad, India.} 
\thanks{{\tt\small\{kaustab21, meduri99aditya\}@gmail.com, mkrishna@iiit.ac.in}}%
\thanks{$^2$ is with IIT Jodhpur, India. {\tt\small avinashsharma@iitj.ac.in}}%
\thanks{Codebase is available at \url{https://tinyurl.com/atppnet}}
}

\begin{document}

\maketitle
\thispagestyle{empty}
\pagestyle{empty}

%%%%%%%%%%%%%%%%%%%%%%%%%%%%%%%%%%%%%%%%%%%%%%%%%%%%%%%%%%%%%%%%%%%%%%%%%%%%%%%%
\begin{abstract}
Point cloud prediction is an important yet challenging task in the field of autonomous driving. The goal is to predict future point cloud sequences that maintain object structures while accurately representing their temporal motion. These predicted point clouds help in other subsequent tasks like object trajectory estimation for collision avoidance or estimating locations with the least odometry drift. In this work, we present ATPPNet, a novel architecture that predicts future point cloud sequences given a sequence of previous time step point clouds obtained with LiDAR sensor. ATPPNet leverages Conv-LSTM
 along with channel-wise and spatial attention
 dually complemented by a $3D$-CNN branch for extracting an enhanced spatio-temporal context to recover high quality fidel predictions of future point clouds. 
We conduct extensive experiments on publicly available datasets and report impressive performance outperforming the existing methods. We also conduct a  thorough ablative study of the proposed architecture and provide an application study that highlights the potential of our model for tasks like odometry estimation.
\end{abstract}
% \Madhav{Abstract is not numbered this way}

% \Madhav{How about the title as ATPPNet}

%%%%%%%%%%%%%%%%%%%%%%%%%%%%%%%%%%%%%%%%%%%%%%%%%%%%%%%%%%%%%%%%%%%%%%%%%%%%%%%%
\section{Introduction}
Autonomous navigation is a widely explored research direction in the robotics domain with applications in autonomous aerial/aquatic drones, vehicles, mobile robots, etc. Recent advancements in $3D$ sensing led by commercial LiDAR (Light Detection and Ranging) technology have reinvigorated interest in this field as LiDAR sensors yield large-scale real-time sequential point clouds ( also represented as range images), providing high-fidelity perception of the $3D$ world in comparison to traditional monocular/stereo based vision solutions. The availability of such large-scale data \cite{caesar2020nuscenes, kitti} has enabled researchers to explore relevant complex tasks such as Localization \cite{chen2021range, shan2018lego}, Place Recognition \cite{ma2022seqot}, Segmentation \cite{chen2021moving, milioto2019rangenet++} and Obstacle Trajectory Prediction \cite{luo2018fast}. The majority of existing methods attempting to solve these tasks rely on captured sequential point clouds available in a given temporal window of the recent past. Interestingly, predicting the future $3D$ point cloud that the sensor is likely to see, can immensely enhance the performance of autonomous navigation tasks like active localization \cite{omama2022drift}. 

Nevertheless,  the task of predicting future point clouds comes with its own set of challenges. One key challenge is that the point clouds are unordered in the space dimension (albeit ordered temporally) and vary in sampling size hence it is difficult to model spatio-temporal coherence among them. As a result, conventional architectures for feature encoding (e.g., CNNs) and sequence prediction 
 (e.g., LSTMs) cannot be directly employed as they cannot process spatially unordered data. Another key challenge is that the LiDAR point clouds are extremely sparse making it difficult to capture the geometrical structures of the objects in the scene and hence predicting them in the future timesteps is extremely difficult. The noise in sensing puts additional challenges in the perception of real-world scenes where objects are largely cluttered. 
Moreover, each full-scale point cloud contains more than $100,000$ points. Extracting features from these sequences of full-scale point clouds becomes a memory-intensive task.  

\begin{figure}[t]
    \centering
    \includegraphics[width=\linewidth]{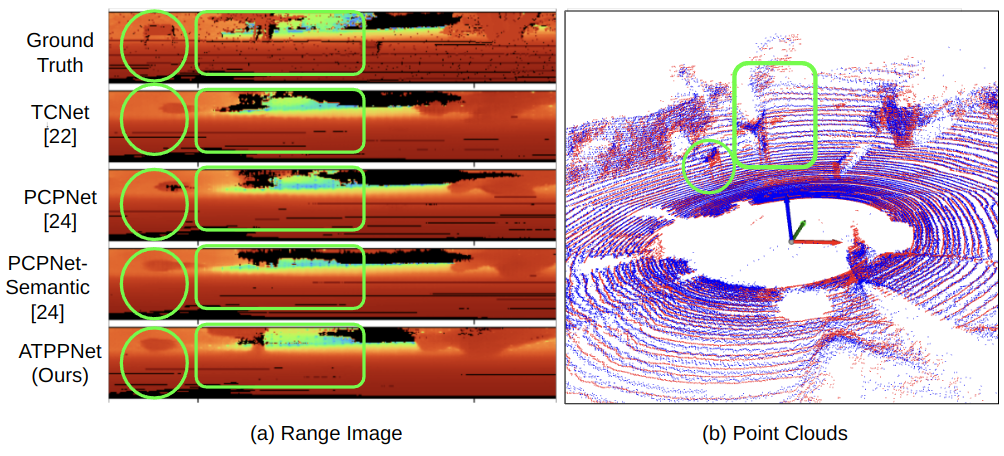}
    \caption{(a) Predicted range images by our ATPPNet and existing methods in comparison to ground truth and, (b) the  $3D$ rendering of the predicted point cloud by ATPPNet (blue) and ground-truth (red). Green circle/rectangle highlights regions where ATPPNet's predictions are superior. 
    % The area marked with the green circle shows the prediction of a sign board and the area marked with the green rectangle shows the prediction of the tree trunk and the canopy. We can observe that ATPPNet is qualitatively closest to the ground truth range image and the point cloud. 
    } \label{fig:teaser}
\end{figure}

Traditionally, $3D$ data is processed with deep learning encoders using volumetric~\cite{maturana2015voxnet, riegler2017octnet, wang2017cnn}, point-cloud~\cite{qi2017pointnet, yan2020pointasnl} and multi-view projection~\cite{su15mvcnn, yu2018multi, yang2019learning} methods. 
 In regard to future point cloud prediction, primarily two lines of work exist, focusing on point cloud and range image representation. The existing point cloud prediction methods either reformulate the task as scene flow estimation~\cite{deng2020temporal} or employ RNN kind of temporal prediction \cite{fan2019pointrnn,lu2021monet}. 
% 3D motion field prediction is also well attempted in literature~\cite{liu2019flownet3d, yuan20203dmotion,aksan2021spatio}.
 The former predicts just a translation of the $3D$ points and hence does not represent the future point cloud accurately. At the same time, the latter works on down-sampled point clouds (for memory efficiency reasons) thereby limiting the resolution of $3D$ data. 
 
 On the other hand, range image based representations project the point cloud data to a $2D$ virtual image plane of the LiDAR sensor, thereby retaining only the single (closest, farthest, or average) depth of the scene for every pixel. Early work with this representation \cite{spf2} used LSTMs to process the temporal sequences and predict a sequence of future range images. 
 %The LSTMs are good at modelling temporal sequences, they are not good at modelling the spatial structures. 
\cite{mersch2022self} used $3D$-CNNs with circular padding and skip-connections to predict a sequence of future range images while \cite{weng2022s2net} used Conv-LSTMs on each of the features from the convolution encoder for the prediction task. However, their network is cumbersome and they use the auto-regressive approach for prediction of range images. Recent work in~\cite{pcpnet} uses the self-attention mechanism of Transformers along with a semantic-based loss function. This method compresses the $3D$ tensors into height and width dimensions and processes each of them separately using two separate transformer blocks. As a result, they are using self-attention only on the channels and since they are compressing the feature tensor into height and width they are also losing the spatial context. Additionally, their model size in terms of the number of parameters is large.

In this paper, we propose a novel architecture for predicting future point clouds from a given sequence of past point clouds represented as LiDAR range images. More specifically, we propose {\it ATPPNet: Attention based Temporal Point cloud Prediction Network} that leverages Conv-LSTM \cite{shi2015convolutional} blocks along with channel-wise and spatial attention modules for extracting an enhanced spatio-temporal context for the task of future point cloud prediction. Further, we also leverage a complimentary $3D$-CNN branch to spatio-temporally encode the global feature embeddings of the range images. Additionally, we also predict the re-projection mask associated with the predicted range images to retain only the valid range values when re-projecting to the point cloud.
%Convolutional LSTM blocks cannot process unordered data, so we project the point clouds onto the spherical coordinate system and then represent them as $2D$ range images. 
Compared to \cite{pcpnet}, 
we show that processing the range image sequences using Conv-LSTM and using spatial and channel-wise attention directly on learned spatio-temporal $3D$ features works better without the need for a separate semantic-based loss function. Our proposed architecture achieves state-of-the-art performance on two publicly available datasets. Our method yields real-time future point cloud prediction (faster than a typical rotating $3D$ LiDAR sensor point cloud rate i.e., 10Hz). We conduct thorough qualitative and quantitative evaluations as well as provide a detailed ablation study to validate the effectiveness of our proposed architecture. %All our claims are backed by the paper thorugh experimental evaluations.
To summarize, our main contributions are as follows:
\begin{itemize}
    \item We proposed a novel architecture (ATPPNet) that leverages Conv-LSTM and spatial and channel-wise attention for predicting future point clouds from a sequence of past point clouds.
    \item ATPPNet achieves SOTA performance on various publicly available datasets while beating the existing methods by $8-10 \%$ margin.
  %  \item Extensive experiments are performed to validate the effectiveness of our method, and in-depth analysis is provided to illustrate the motivations behind our design.
    \item We empirically show that ATPPNet improves the performance of downstream tasks, like odometry estimation. %thus verifying that our predictions are better suited for active localization tasks. % where we need to predict how our localization will be
    % \item We also showcase the application of predicting ATPPNet i
\end{itemize}

% \Madhav{Typically we summarize the Into with an enumeration of the contributions and linking to the Tables, Plots in the Results Section to the Corresponding Contribution}

%%%%%%%%%%%%%%%%%%%%%%%%%%%%%%%%%%%%%%%%%%%%%%%%%%%%%%%%%%%%%%%%%%%%%%%%%%%%%%%%
%\input{relatedworks}
%%%%%%%%%%%%%%%%%%%%%%%%%%%%%%%%%%%%%%%%%%%%%%%%%%%%%%%%%%%%%%%%%%%%%%%%%%%%%%%%
\section{Our Approach} \label{architecture}
We provide details of our novel ATPPNet (Attention based Temporal Point cloud Prediction Network) that leverages Conv-LSTM blocks
along with channel-wise and spatial attention modules complemented by a 3D-CNN branch to process a past sequence of $3D$ point clouds to predict future point clouds.
\begin{figure*}[h!]
    \centering
    \begin{subfigure}{\textwidth}
        \includegraphics[width=\textwidth]{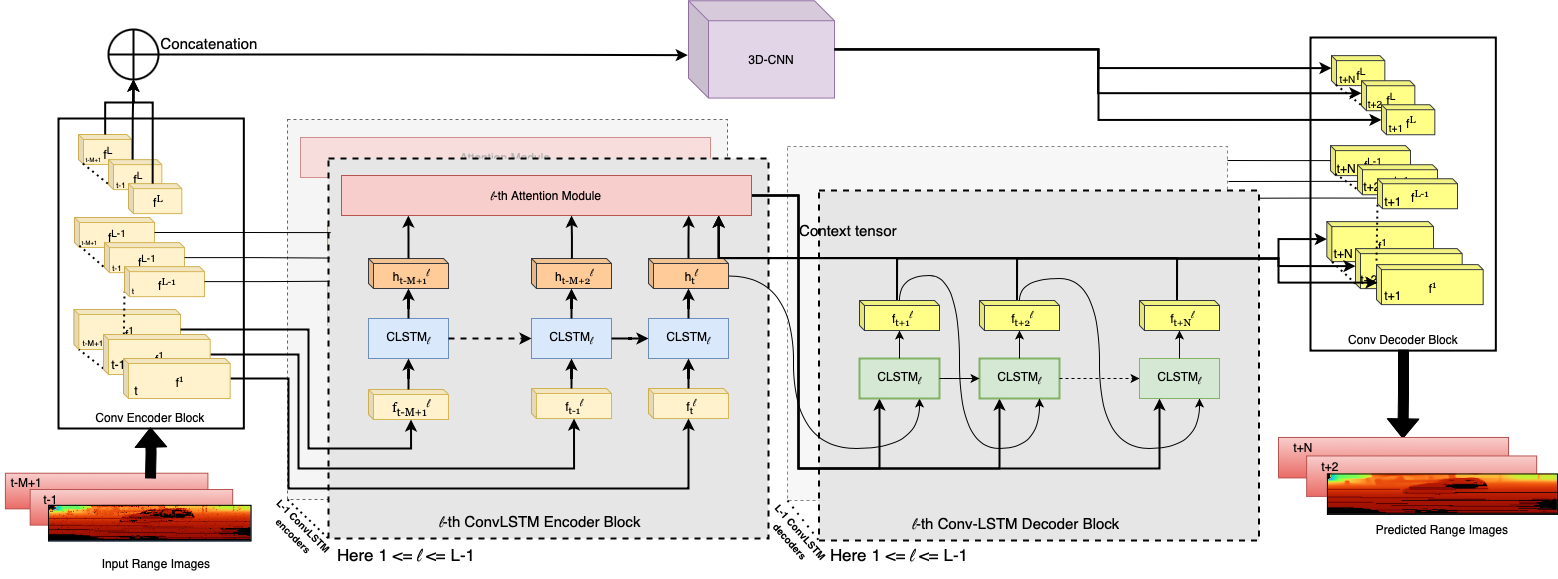}
    \end{subfigure}
    \caption{\textbf{ATPPNet Architecture.} ATPPNet leverages Conv-LSTM along with channel-wise and spatial attention dually complemented by a 3D-CNN branch for extracting an enhanced spatio-temporal context to recover high quality fidel predictions of future point clouds.} 
%     We leverage Conv-LSTMs along with channel-wise and spatial attention modules to process
% past sequences of point clouds and predict future sequences. We also leverage complimentary $3D$-CNN to spatio-temporally encode the global feature embeddings of the range image.}
    % $L$ feature tensors from the input sequence of range images are first extracted using the CNN encoder block. Each of the sequence of the $L$ feature tensors are then modelled spatio-temporally using the Conv-LSTM block. The output features from the Conv-LSTM blocks are then refined using the spatial and channel-wise attention module and a context tensor is generated for eaach of the $L$ sequence of features. These context tensors are then used by the decoder Conv-LSTM block to generate the feature tensors which are then up-sampled using the convolutional decoder block to generate the sequence of range images.  }
    \label{fig:architecture_overview}
\end{figure*}
%
% \Madhav{It is very important to have the architecture figure that we refer to while explaining}
%
%A pointcloud sequence for $T$ time steps is denoted as $S = \{S_1, S_2, \dots, S_T\}$. The goal of our task is to predict the future sequence of full-scale point clouds denoted as $S_F = \{S_{t+1}, S_{t+2}, \dots, S_{t+N}\}$, given a past sequence denoted as $S_P = \{S_{t-M+1}, S_{t-M+2}, \dots S_t\}$, where $N$ and $M$ are the number of future and past time steps respectively.
Let $S_P = \{S_{t-M+1}, S_{t-M+2}, \dots S_t\}$ be the set of input $3D$ point cloud sequence of $M$ time steps in a temporal window where $S_{\tau} \in \mathbb{R}^3$ is set of $3D$ points captured at specific time step $\tau$. The goal of our ATPPNet is to predict the future sequence of $3D$ point clouds for a temporal window of $N$ time steps represented as $S_F = \{S_{t+1}, S_{t+2}, \dots, S_{t+N}\}$. Similar to~\cite{mersch2022self}, we adopt the range image representation by first converting the point clouds into the spherical coordinate system and then projecting the corresponding point cloud ($S_{\tau} \in \mathbb{R}^3$) to the virtual image plane of the LiDAR sensor, represented as $R_{\tau} \in \mathbb{R}^{2}$. 
Let $R_P = \{R_{t-M+1}, R_{t-M+2}, \dots R_t\}$ be the sequence of past range images in a fixed temporal window (obtained from $S_P$ ) and similarly $R_F = \{R_{t+1}, R_{t+2}, \dots, R_{t+N}\}$ be the sequence of predicted range images associated with $S_F$.  
% 3D scans.

%
%TODO: Move to Intro Each full-scale pointclouds contains more than $100,000$ points. As a result, extracting features from a sequence of full-scale pointcloud becomes a memory intensive task and increases the space complexity of the designed system by a large margin. To keep our system efficient in terms of memory, we use $2D$ representations of the pointclouds. While there are different kinds of $2D$ representations, we followed the approach used by the previous works \cite{mersch2022self, weng2022s2net, pcpnet} and represented the point clouds as range images by first converting the point clouds into the spherical coordinate system and then projecting it onto the $2D$ image plane of the LiDAR sensor. 
\subsection{Overall Architecture}

%Our proposed ATPPNet use a ConvLSTM based architecture with spatial and channel-wise attention module. 
Figure~\ref{fig:architecture_overview} provides the overview of the proposed ATPPNet architecture. A shared convolution encoder processes the input range images $R_P$ and generates $L$ number of multi-scale feature tensors for each of the range images. Subsequently, the first $L-1$ feature tensors are fed to Conv-LSTMs to model the spatio-temporal relationships across $R_P$. 
Further, we exploit spatial as well as channel-wise attention on the outputs of Conv-LSTMs to obtain the $L-1$ {\it context tensors} (i.e., the consolidated spatio-temporal encoding of $R_P$). 
Additionally, for the final $L$-th feature tensor, we use a $3D$-CNN layer to process the spatio-temporal relationship and generate the $L$-th feature tensor for $N$ future time steps.
% Using conv-LSTMs for each of the $L-1$ layers acts similar to skip-connections and helps preserve the high-frequency details. 
On the decoder side, we feed the context tensor along with the hidden state of the last time step to $L-1$ Conv-LSTMs and generate the feature tensors for each of the $L-1$ layers for $N$ time steps into the future. All these $L$ feature tensors on the decoder side are subsequently processed to generate the range image sequence $R_F$ along with their corresponding re-projection masks $M_F$ for all the $N$ future time steps where each pixel of $M_{\tau} \in M_{F}$ can be interpreted as the probability for each of the range image pixels to be valid or invalid. This re-projection mask is used while back-projecting a range image to a point cloud where we only retain the range values corresponding to probabilities greater than $0.5$.
The construction of specific architectural blocks is given below. 
% For predicting the range image for time step $n$ where $t+1 \leq n \leq N$, the decoder conv-LSTMs takes as input a context feature along with the hidden state for the $n-1$ conv-LSTM.

\subsection{Convolution Encoder \& Decoder block} \label{conv_encoder_decoder}
% \subsection{Convolution Encoder} \label{conv_encoder_decoder}
%
The convolution encoder block takes the range image and first performs a $2D$-convolution operation, 
% followed by $2D$-batch normalization and leaky-ReLU
resulting in a tensor with an increased number of channels but the same spatial resolution.
% of dimension $C \times H \times W$. 
This tensor is further processed using $L$ convolutional sub-blocks.
% Adi - The convolution encoder block takes the range image (of dimension $1 \times H \times W$) and first performs a $2D$-convolution operation, resulting in a tensor of dimension $C \times H \times W$. This tensor is further processed using $L$ convolutional blocks.
% The Convolution encoder block consists of $L$ sub-blocks. 
Each sub-block takes as input a feature tensor 
% of dimension $C_{l-1} \times H_{l-1} \times W_{l-1}$ 
and applies a combination of $2D$-convolution, $2D$-batch normalization, and leaky-ReLU operation while keeping the tensor dimensions the same. 
A strided-convolution operation is subsequently performed, resulting in a down-sampled tensor.
The convolution decoder block follows the reversed structure of the convolution encoder block. There are $L$ sub-blocks, each of which takes an input tensor
% of dimension $C_{l-1} \times \frac{H_{l}}{h_l} \times \frac{W_{l}}{w_l}$ 
and passes it through a $2D$-transposed convolution, $2D$-batch normalization, and leaky-ReLU operation resulting in a spatially scaled-up tensor while keeping the number of channels the same.
% while keeping the channel size same.
% to a dimension of $C_{l-1} \times H_{l} \times W_{l}$. 
% After performing a 2D batch normalization and leaky-ReLU operation on this tensor, 
Another $2D$-convolution operation is then performed, to decrease the channel size of the tensor.
% and again perform a batch norm and leaky-ReLU on this.
The output of the $L$-th layer is finally passed through another $2D$-convolutional layer resulting in the predicted range images and the associated re-projection mask.

\subsection{Conv-LSTM encoder}

%The $L$ feature tensors obtained with the convolutional encoder block encode different features (shapes, edges, etc.) of the content in the range images. Since we are dealing with a sequence of $M$ range images, for each of the $L$ feature tensors we will have $M$ sequences. 
We propose to use $L-1$ Conv-LSTMs to exploit the spatio-temporal context of input sequences. Similar to the S2Net \cite{weng2022s2net} architecture, we used multiple Conv-LSTM layers for each of the feature tensors.
% to acts as temporally aligned skip connections \cite{weng2022s2net}. 
This helps us in preserving the high-frequency details across the range image sequences. 
The Conv-LSTMs for each of the $L-1$ features generate a hidden state for each of the $M$ time steps. %These hidden-states are processed further using the Attention Module ( see \autoref{attention}) to generate the context tensor to be used by the decoder CLSTMs.

\subsection{Attention Module} \label{attention}
Let $\tau \in [t-M+1, \dots, t-1, t]$ be a specific time step in the given input temporal window, and the feature tensors for the layer $l$ at every time step be represented as $f_{\tau}^l$. Similarly, let the output of $l$-th Conv-LSTM at time step $t$ denoted as $g_t^l$ where $l \in [1, \dots L-1]$. As part of the attention module, we first compute the joint embedding $\mathcal{J}_{\tau}^l$ of $f_{\tau}^l$ and $g_t^l$ as (using the formulation \cite{schlemper2018attention}):

$$
\mathcal{J}_{\tau}^l = \sigma_1(W_f f_{\tau}^l \oplus W_g g_t^l) ,
$$
where $\sigma_1$ is a non-linear activation function chosen as ReLU and $\oplus$ is the concatenation operation. $W_f$ and $W_g$ are implemented as $2D$-convolution operations with $1 \times 1$ kernel. The use of $\sigma_1$, $W_f$, and $W_g$ allows the network to learn non-linear relationships between the features, which is especially important when the image is noisy like our range images. The resulting 
tensors $\mathcal{J}_{\tau}^l$ are
% tensors $\mathcal{J}_{\tau}^l$ is a 3D-tensor that is 
passed through the spatial and channel-wise attention module \cite{park2018bam} to find the $3D$ attention map $\mathcal{M}(\mathcal{J}_{\tau}^l) \in R^{C_l \times H_l \times W_l}$. The refined feature tensor for layer $l$ at time step $\tau$ is computed as:
$$
\hat{f}_{\tau}^l = f_{\tau}^l \otimes \mathcal{M}(\mathcal{J}_{\tau}^l).
$$

Here $\otimes$ is the element-wise multiplication. To compute the $3D$ attention map $\mathcal{M}(J_{\tau}^l)$, we compute the channel-wise attention and spatial attention separately and then combine them as

$$
\mathcal{M}(\mathcal{J}_{\tau}^l) = \sigma (M_c(\mathcal{J}_{\tau}^l) \otimes M_s(\mathcal{J}_{\tau}^l))a,
$$

where $\sigma$ is the Sigmoid function and $\otimes$ is the element-wise multiplication operation. The $M_c$ function first applies the global average pooling operation on the $3D$ tensor $\mathcal{J}_{\tau}^l$ to get the channel tensor which is then passed through an MLP layer to get the channel-wise attention values. The $M_s$ function applies $2D$-convolution operation on the $3D$ tensor $\mathcal{J}_{\tau}^l$ and returns a single channel tensor which represents the spatial attention values.

%We chose to go with sigmoid instead of softmax because the softmax operation was producing extremely sparse attention values which resulted in poor performance.

The refined feature tensors for all the $M$ time steps for each of the $L-1$ layers are then used to compute the context tensors, that are subsequently served as input to the decoder Conv-LSTMs.

$$
context_l^{t} = \sum_{\tau = t}^{t-M+1} \hat{f}_{\tau}^l
$$
% \Madhav{What are the $\otimes$ and $\oplus$ operators need to be mentioned}

\subsection{Conv-LSTM decoder}
The Conv-LSTM decoder follows a similar structure as the Conv-LSTM encoder.
$L-1$ Conv-LSTM decoders are used to predict $L-1$ feature tensors for each of the $N$ future time steps. 
% These features are then used by the convolution decoder block to generate the predicted range image. 
Let $\tau \in [t+1, \dots, N]$ be a specific time step in the predicted future temporal window.
For the $\tau$-th time step, the $l$-th Conv-LSTM decoder takes as input the context tensor $context_l^{\tau -1}$ where $l \in [1, \dots, L-1]$ along with the hidden state of the Conv-LSTM for time step $\tau-1$ to compute the output feature tensor.
% produces an output tensor on the decoder side. 
This output feature tensor along with the hidden states of the previous time steps are used to re-compute the context tensor $context_l^{\tau}$ to be used for the next time step. 
These output feature tensors on the Conv-LSTM decoder side are used by the convolutional decoder to generate the predicted range images $R_F$

\subsection{$3D$-CNN block}

% The top-most layers of a convolution encoder learns to represent the high-frequency details of the range image whereas the lower layers tend to capture the local representations \cite{zeiler2014visualizing}, \cite{yosinski2015understanding}. 
The feature tensors for the $L^{th}$ layer from the convolutional encoder block for all the past $M$ time steps are concatenated to create a $4D$ tensor.
% of shape $M \times C_L \times H_L\times W_L$. 
A $3D$-CNN layer is used to process this feature tensor and generate $N$ feature maps for the $L$-th convolutional decoder block. 

It is important to note that since $3D$-CNNs place their spatial focus on fewer, contiguous areas in the feature tensors \cite{manttari2020interpreting}, we employ $3D$-CNN on the last $L-th$ layer of the feature tensor (obtained with $2D$-CNN) as it tends to capture global structures in the range image \cite{zeiler2014visualizing}, \cite{yosinski2015understanding}. Thus, the $3D$-CNN block extracts only the complementary spatio-temporal context as the primary spatio-temporal context is already obtained by applying Conv-LSTM's on the initial layers of the convolutional encodings as they tend to capture the high frequency details in the range images. %it improves our performance over using Conv-LSTM to model the $L$-th feature tensor. 
This also gives us the additional advantage of speeding up our inference time.
% Conv-LSTMs have slower inference time and as a result using $L$ Conv-LSTM blocks, one block per feature tensor of the input range image, to predict $R_F$ might lead to a longer inference time. To keep the inference time below the operating frequency of the LiDAR, we replace the $L^{th}$ Conv-LSTM layer with a $3D$-CNN layer. The feature tensors for the $L^{th}$ layer from the CNN Encoder block for all the past $M$ time steps are concatenated to create a $3D$ tensor of shape $M \times C_L \times H_L\times W_L$. This $3D$ tensor is then processed using the 3D-CNN layer to generate $M$ feature maps for the $L$-th convolutional decoder sub-block.

\subsection{Loss Function}

We use a combination of losses when training the network.
% When training the network, we used a combination of multiple losses. 
Since our ground truth point clouds are projected onto $2D$ range images of dimension $H \times W$, we can use $2D$ image-based losses. 
% Let $R_F$ be the predicted range images for $F$ time steps into the future.

Firstly, we use the average range loss $\mathcal{L}_R$ to compute the error between the predicted range values $\hat{r}_{\tau,i,j} \in \mathbb{R}^{N \times H \times W}$ and the ground-truth range values $r_{\tau,i,j} \in \mathbb{R}^{N \times H \times W}$. The average range loss can be formulated as: 
$$
\mathcal{L}_R = \frac{1}{N \times H \times W} \sum_{\tau = t+1}^{t+N}\sum_{i=1}^{H}\sum_{j=1}^{W} \mid \mid \hat{r}_{\tau,i,j} - r_{\tau,i,j} \mid \mid_1,
$$
where $\mid \mid \bullet \mid \mid_1$ represents the $L_1$ norm. The range loss $\mathcal{L}_R$ is computed only for the valid ground truth points.
To train the re-projection mask output, we use the Binary Cross-Entropy loss between the predicted mask values $\hat{m}_{\tau,i,j} \in \mathbb{R}^{N \times H \times W}$ and the ground truth mask values $m_{\tau,i,j} \in \mathbb{R}^{N \times H \times W}$. The average mask loss $\mathcal{L}_M$ is computed as:

\begin{align}
\mathcal{L}_M &= \frac{1}{N \times H \times W} \sum_{\tau = t+1}^{t+N}\sum_{i=1}^{H}\sum_{j=1}^{W} -m_{\tau,i,j} \log \hat{m}_{\tau,i,j} \\
&- (1-m_{\tau,i,j}) \log(1-\hat{m}_{\tau,i,j}),
\end{align}
where $\hat{m}_{\tau,i,j}$ is the predicted probability of whether the range value is valid. $m_{\tau,i,j}$ is $1$ if the ground-truth range value is valid and $0$ otherwise.
A masked range image is generated by taking only the range values from the range image whose corresponding mask values are greater than $0.5$. Since we are re-projecting the predicted masked range images into point clouds, we use Chamfer distance \cite{pointset} represented as $\mathcal{L}_C$ for evaluating fidelity of the predicted point clouds.
% need to compute a point-based loss to improve the point cloud prediction. Chamfer distance \cite{pointset} is a common metric for evaluating $3D$ point clouds.
% The Chamfer distance loss $\mathcal{L}_C$ is computed as 
% \begin{align}
% \mathcal{L}_C = \frac{1}{N} \sum_{\hat{p}\in \hat{S}} \min_{p \in S} \mid \mid \hat{p} - p \mid \mid^{2}_{2} + \frac{1}{M} \sum_{p\in S} \min_{\hat{p} \in \hat{S}} \mid \mid p - \hat{p} \mid \mid^{2}_{2},
% \end{align}
% where $\mid \mid \bullet \mid \mid_2$ represents the $L_2$ norm, $S$ represents the ground-truth point cloud and $\hat{S}$ represents the predicted point cloud.

The combined loss function is given as 
$$
\mathcal{L} = \mathcal{L}_R + \mathcal{L}_{M} + \alpha_C \mathcal{L}_C.
$$
$\alpha_C$ is the weight associated with the Chamfer distance.
%%%%%%%%%%%%%%%%%%%%%%%%%%%%%%%%%%%%%%%%%%%%%%%%%%%%%%%%%%%%%%%%%%%%%%%%%%%%%%%%

\begin{figure*}[t]
    \centering
    \includegraphics[width=\textwidth]{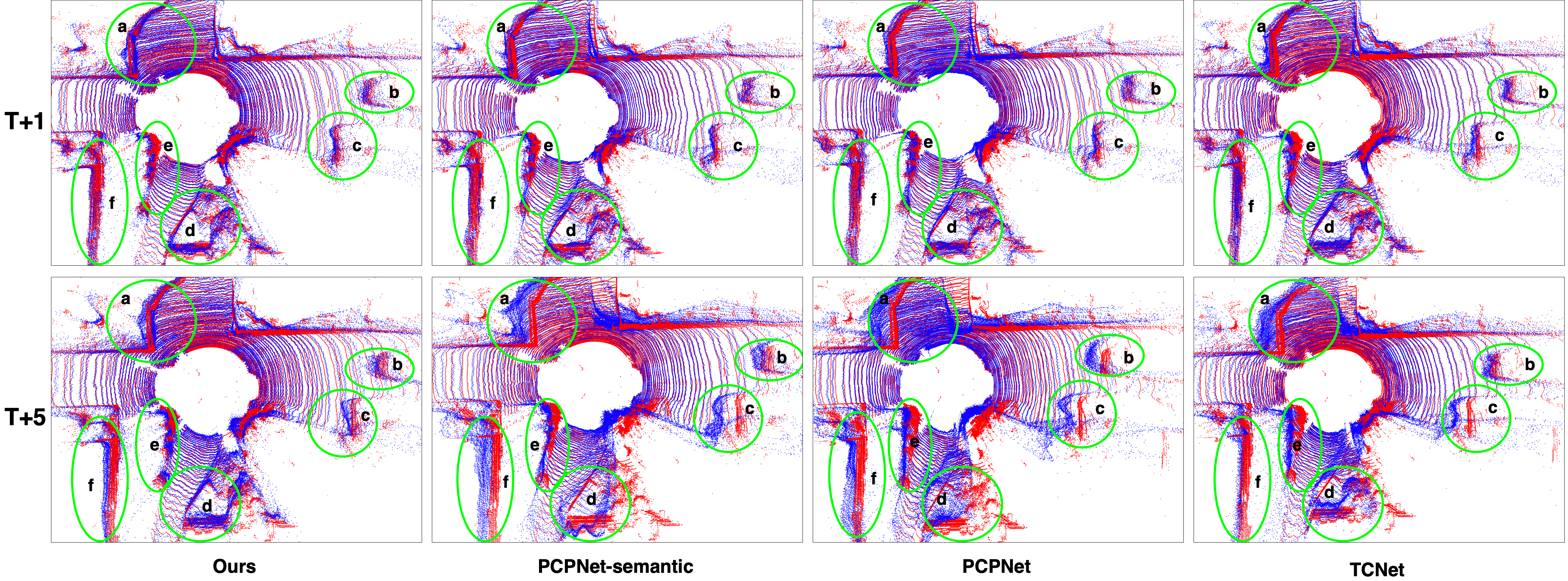}
    \caption{Qualitative comparison conducted on sequence 10 of the KITTI odometry dataset. The predicted points (blue) and the ground truth points (red) are combined for a better visual comparison. The top row shows the point clouds at prediction step $t+1$ and the bottom row shows the point clouds at prediction step $t+5$. The areas of interest are circled in green.} \label{fig:qualitative}
\end{figure*}

\begin{table}[ht]
\begin{center}
\begin{tabularx}{\linewidth}{|>{\centering\arraybackslash}X |>{\centering\arraybackslash}X |>{\centering\arraybackslash}X | >{\centering\arraybackslash}X | >{\centering\arraybackslash}X | >{\centering\arraybackslash}X |} 
 \hline
 %  & \multicolumn{6}{c||}{\textbf{Sampled Point Cloud}} & \multicolumn{4}{c|}{\textbf{Full-Scale Point Clouds}}\\ %[0.5ex] 
 % \hline
 Prediction Step & TCNet \cite{mersch2022self} & PCPNet \cite{pcpnet} & PCPNet-Semantic \cite{pcpnet} & ATPPNet (Ours)\\ [0.5ex] 
 \hline\hline
 1 & 0.554 & 0.543 & 0.503 & \textbf{0.468}\\
 \hline
 2 & 0.671 & 0.662 & 0.620 & \textbf{0.570}\\
 \hline
 3 & 0.779 & 0.773 & 0.727 & \textbf{0.667}\\
 \hline
 4  & 0.878 & 0.872 & 0.825 & \textbf{0.760} \\ 
 \hline
 5  & 0.974 & 0.973 & 0.920 & \textbf{0.851} \\ 
 \hline \hline
 Mean  & 0.771 & 0.765 & 0.719 & \textbf{0.663} \\ 
 \hline
\end{tabularx}
\caption{Range Loss results on the KITTI odometry test set verifies that ATPPNet has a performance improvement of $4.026 \% $ over SOTA on the mean range loss. Bold values correspond to the best performing model in that corresponding time step.}
\label{table:range_loss_comparison}
\end{center}
\end{table}

\begin{table*}[ht]
\begin{center}
\begin{tabularx}{\textwidth}{|>{\centering\arraybackslash}X |>{\centering\arraybackslash}X |>{\centering\arraybackslash}X | >{\centering\arraybackslash}X | >{\centering\arraybackslash}X | >{\centering\arraybackslash}X | >{\centering\arraybackslash}X||>{\centering\arraybackslash}X|>{\centering\arraybackslash}X|>{\centering\arraybackslash}X|>{\centering\arraybackslash}X|} 
 \hline
  & \multicolumn{6}{c||}{\textbf{Sampled Point Cloud}} & \multicolumn{4}{c|}{\textbf{Full-Scale Point Clouds}}\\ %[0.5ex] 
 \hline
 Prediction Step & PointLSTM \cite{fan2019pointrnn} & MoNet \cite{lu2021monet} & TCNet \cite{mersch2022self} & PCPNet \cite{pcpnet} & PCPNet-Semantic \cite{pcpnet} & ATPPNet (Ours) & TCNet \cite{mersch2022self} & PCPNet \cite{pcpnet} & PCPNet-Semantic \cite{pcpnet} & ATPPNet (Ours) \\ [0.5ex] 
 \hline\hline
 1 & 0.332 & 0.278 & 0.290 & 0.285 & 0.280 & \textbf{0.258} & 0.253 & 0.252 & 0.242 & \textbf{0.225}\\
 \hline
 2 & 0.561 & 0.409 & 0.357 & 0.341 & 0.340 & \textbf{0.311} & 0.309 & 0.301 & 0.298 & \textbf{0.270}\\
 \hline
 3 & 0.810 & 0.549 & 0.441 & 0.411 & 0.412 & \textbf{0.375} & 0.377 & 0.362 & 0.354 & \textbf{0.326}\\
 \hline
 4  & 1.054 & 0.692 & 0.522 & 0.492 & 0.495 & \textbf{0.445} & 0.448 & 0.435 & 0.427 & \textbf{0.391} \\ 
 \hline
 5  & 1.299 & 0.842 & 0.629 & 0.580 & 0.601 & \textbf{0.523} & 0.547 & 0.514 & 0.503 & \textbf{0.461} \\ 
 \hline \hline
 Mean  & 0.811 & 0.554 & 0.448 & 0.422 & 0.426 & \textbf{0.382} & 0.387 & 0.373 & 0.365 & \textbf{0.335} \\ 
 \hline
\end{tabularx}
\caption{Chamfer distance results on KITTI Odometry test sequence with the sampled point clouds on the left and full-scale point clouds on the right. ATPPNet has a performance improvement of $10.328 \% $ over SOTA for the sampled point clouds and $8.219 \% $ over SOTA for the full-scale point clouds. Bold values correspond to the best performing model in that corresponding time step.}
\label{table:chamfer_loss_kitti}
\end{center}
\end{table*}

\begin{table}[ht]
\begin{center}
\begin{tabularx}{\linewidth}{|>{\centering\arraybackslash}X |>{\centering\arraybackslash}X |>{\centering\arraybackslash}X | >{\centering\arraybackslash}X | >{\centering\arraybackslash}X | >{\centering\arraybackslash}X || >{\centering\arraybackslash}X|} 
 \hline
 Evaluation metric & TCNet \cite{mersch2022self} & PCPNet-Semantic \cite{pcpnet} & ATPPNet (Ours) \\ [0.5ex] 
 \hline\hline
 Mean Chamfer Distance & 1.389 & 1.360 & \textbf{0.932}\\
 \hline
 Mean Range Loss & 0.719 & 0.704 & \textbf{0.598}\\
 \hline
\end{tabularx}
\caption{Mean Range Loss and Chamfer distance on the nuScenes test set. ATPPNet is making an improvement of $15.056 \% $ over SOTA on the mean Range loss and $31.470 \% $ over SOTA on the mean Chamfer distance.}
\label{table:nu_scenes_loss_comparison}
\end{center}
\end{table}

% \vspace{-20pt}
\begin{table*}[ht]
\begin{center}
\begin{tabularx}{\textwidth}{|>{\centering\arraybackslash}X |>{\centering\arraybackslash}X |>{\centering\arraybackslash}X | >{\centering\arraybackslash}X || >{\centering\arraybackslash}X | >{\centering\arraybackslash}X | >{\centering\arraybackslash}X||>{\centering\arraybackslash}X|} 
 \hline
  & \multicolumn{3}{c||}{\textbf{A) Attention Module \label{attention}}} & \multicolumn{3}{c||}{\textbf{B) Conv-LSTM (CLSTM) layers}} &\\ %[0.5ex] 
 \hline
 Evaluation metric & No Attention & S-Attention & C-Attention & $L-1$ CLSTM & $L-1$ \& $L-2$ CLSTM & all $L$ CLSTM & ATPPNet (Ours) \\ [0.5ex] 
 \hline\hline
 Chamfer distance & 0.365 & 0.359 & 0.356 & 0.405 & 0.378 & 0.366 & \textbf{0.335} \\
 \hline
 Range loss & 0.719 &  0.687 & 0.690 & 0.770 & 0.698 & 0.717 & \textbf{0.663} \\
 \hline
\end{tabularx}

% \caption{Abblation study on the attention module and the number of temporally aligned CLSTM. The Loss values are computed on the KITTI odometry test sequence $08-10$. Ours use spatial and channel-wise attention and $3$ temporally-aligned CLSTM. Bold values represent the optimal performance. }
\caption{Results of Ablation study on Attention Module and Conv-LSTM layers. Bold values correspond to the best performing model.}
\label{table:abblation}
\end{center}
\end{table*}

\begin{table}[ht]
\begin{center}
\begin{tabularx}{\linewidth}{|>{\centering\arraybackslash}X |>{\centering\arraybackslash}X |>{\centering\arraybackslash}X | >{\centering\arraybackslash}X | >{\centering\arraybackslash}X |} 
 \hline
 Window size & 3 & 5 & 7\\ [0.5ex] 
 \hline\hline
 T+1  & \textbf{0.221} & 0.255 & 0.258\\ 
 \hline
 T+2 & 0.274 & \textbf{0.270} & 0.306\\ 
 \hline
 T+3 & 0.344 & \textbf{0.326} & 0.363\\ 
 \hline
 T+4 & NA & \textbf{0.391} & 0.426\\ 
 \hline
 T+5 & NA & \textbf{0.461} & 0.498\\ 
 \hline
 T+6 & NA & NA & 0.580\\ 
 \hline
 T+7 & NA & NA & 0.657\\ 
 \hline
\end{tabularx}
\caption{An ablation study on how the performance changes as we vary the sequence length.}
\label{table:seq_length}
\end{center}
\end{table}

\begin{table}[ht]
\begin{center}
\begin{tabularx}{\linewidth}{|>{\centering\arraybackslash}X |>{\centering\arraybackslash}X |>{\centering\arraybackslash}X | >{\centering\arraybackslash}X | >{\centering\arraybackslash}X | >{\centering\arraybackslash}X |} 
 \hline
 %  & \multicolumn{6}{c||}{\textbf{Sampled Point Cloud}} & \multicolumn{4}{c|}{\textbf{Full-Scale Point Clouds}}\\ %[0.5ex] 
 % \hline
 Pose error            & TCNet \cite{mersch2022self} & PCPNet \cite{pcpnet}   & PCPNet-Semantic \cite{pcpnet} & ATPPNet (Ours) \\ [0.5ex] 
 \hline\hline
 $\mathcal{L}_P^{t+1}$ & 0.1342 & 0.1363 & 0.1280          & \textbf{0.1209}\\
 \hline
 $\mathcal{L}_P^{t+2}$ & 0.2412 & 0.2282  & 0.2235           & \textbf{0.2065}\\
 \hline
 $\mathcal{L}_P^{t+3}$ & 0.3670 & 0.3388  & 0.3343           & \textbf{0.3038}\\
 \hline
 $\mathcal{L}_P^{t+4}$  & 0.5084 & 0.4630 & 0.4558          & \textbf{0.4128} \\ 
 \hline
 $\mathcal{L}_P^{t+5}$  & 0.6736 & 0.6022 & 0.5878          & \textbf{0.5328} \\ 
 \hline \hline
 Mean                   & 0.3849 & 0.35374 & 0.3459          & \textbf{0.3154} \\ 
 \hline
\end{tabularx}
\caption{\textbf{LOAM pose error.} We adopt LOAM [36] and evaluate the disparity between the motion estimates on ground truth and predictions.}
\label{table:loam_pose_error}
\end{center}
\end{table}

\section{Experiments and Results}
\subsection{Experimental Settings}
We train ATPPNet in a self-supervised manner in the sense that we use only sequential point cloud data sans no manually annotated labels. 
For our experiments, we keep the temporal window size $M = N = 5$. In the convolutional encoder block, the initial convolutional operation outputs $16$ channels while retaining the spatial dimension. We use $L=4$ sub-blocks where the channel size increases by a factor of $2$ for every successive sub-block obtained by the convolutional encoder. In each of the sub-blocks, the first convolutional operation uses a kernel size of $3 \times 3$ with stride $(1,1) $, and the second convolution operation uses a kernel size of $2 \times 4$ with stride $(2, 4) $. All the convolutional operations use circular padding \cite{mersch2022self}.
% We use a kernel size of $(3,3)$ and circular padding
 % use circular padding to maintain spatial consistency on the horizontal borders of the range images
% \cite{mersch2022self} to keep the channel size and spatial dimension of the feature tensor the same, and a kernel size of $(2,4)$ with stride $(2,4)$ and circular padding to upsample and downsample the feature tensors. 
Each Conv-LSTM block uses $3$ layers.
% where the hidden vector size of the $L$-th Conv-LSTM block is similar to the size of the feature tensor from the $L$-th convolutional sub-block. 

Similar to the trend in the literature \cite{mersch2022self, pcpnet}, we train our architecture for $50$ epochs with $\alpha_C = 0$ and then fine-tuned with the Chamfer distance loss for the next $10$ epochs by setting $\alpha_C = 1$. We train our model on a system with an Intel Xeon E5-2640 CPU and $3$ Nvidia RTX 2080 GPUs using the Distributed Data Parallel strategy. While training, we have used the ADAM optimizer \cite{kingma2014adam} with default parameters and an initial learning rate of $0.0003$ and the StepLR learning rate scheduler with gamma as $0.99$.
\subsubsection{KITTI Odometry dataset \cite{kitti}}
We use sequences $00-05$ for training, $06-07$ for validation, and $08-10$ for testing. The LiDAR used in the KITTI dataset \cite{kitti} has $64$ channels, so we have used range images of size $64 \times 2048$.
\subsubsection{nuScenes dataset \cite{caesar2020nuscenes}}
We trained our network on this dataset with the same training strategy we used on the KITTI dataset. Following PCPNet \cite{pcpnet}, we used sequence $00-69$ for training, scenes $70-84$ for validation, and $85-99$ for testing. We trained our architecture on range images of size $32 \times 1024$ since the LiDAR used here has $32$ channels.

\subsection{Qualitative Analysis}

\autoref{fig:qualitative} (and~\autoref{fig:teaser}) shows a qualitative comparison of the predicted point clouds generated using our proposed ATPPNet and the other methods: TCNet \cite{mersch2022self}, PCPNet and PCPNet-semantic \cite{pcpnet}. The areas of interest are highlighted with numbered green circles.

In the predicted sequence $t+1$ shown in ~\autoref{fig:qualitative}, we can observe over the circles $a,b,c$ \& $d$ that our ATPPNet outperforms the other methods by generating point clouds that are less noisy and structurally more similar to the ground truth. 
We can observe in time step $t+5$ (bottom row) that the circles numbered $a,b,c,d,e$ \& $f$ in the point cloud generated by ATPPNet is more fidel to the ground truth compared to the predicted point clouds from the other methods that have large visible deviations from the ground truth and are more noisy.

\subsection{Quantitative Analysis}\label{analysis}

In this section, we perform a quantitative analysis of our proposed ATPPNet with two point based methods (PointLSTM \cite{fan2019pointrnn}, MoNet \cite{lu2021monet}) on the KITTI  \cite{kitti} test set, and three range image based methods (TCNet \cite{mersch2022self}, PCPNet and PCPNet-semantic \cite{pcpnet}) on the KITTI  \cite{kitti} and the nuScenes \cite{caesar2020nuscenes} test set. 
% We also evaluate ATPPNet with the range image based methods on the nuScenes \cite{caesar2020nuscenes} test-set. 
% In robotics applications, it is often more efficient in terms of speed and memory to perform tasks on down-sampled point clouds. 
The point based methods \cite{fan2019pointrnn,lu2021monet} use down-sampled point clouds to $65536$ points and this is also adopted by us and other methods i.e.,  \cite{mersch2022self, pcpnet}.
We use the range loss and Chamfer distance to evaluate the predicted range images and the point clouds, respectively.

% \begin{table}[ht]
% \begin{center}
% \begin{tabularx}{\linewidth}{|>{\centering\arraybackslash}X |>{\centering\arraybackslash}X |>{\centering\arraybackslash}X | >{\centering\arraybackslash}X | >{\centering\arraybackslash}X | >{\centering\arraybackslash}X || >{\centering\arraybackslash}X|} 
%  \hline
%  Method & T+1 & T+2 & T+3 & T+4 & T+5 & Mean\\ [0.5ex] 
%  \hline\hline
%  TCNet & 0.5543 & 0.6711 & 0.7799 & 0.8782 & 0.9747 & 0.7716 \\
%  \hline
%  PCPNet & 0.54380 & 0.6626 & 0.7736 & 0.8727 & 0.9736 & 0.7652\\
%  \hline
%  PCPNet-semantic & 0.5037 & 0.6205 & 0.7279 & 0.8257 & 0.9204 & 0.7197\\
%  \hline
%  Ours  & \textbf{0.4685} & \textbf{0.5706} & \textbf{0.6673} & \textbf{0.7600} & \textbf{0.8518} & \textbf{0.6636} \\ 
%  \hline
% \end{tabularx}
% \caption{Range Loss results on the KITTI odometry test sequences $08-10$. Bold values represent the optimal option in that corresponding time step.}
% \label{table:l1_loss_comparison}
% \end{center}
% \end{table}

\autoref{table:range_loss_comparison} shows the quantitative results of the range loss for all the methods on the KITTI test set. Compared to the other methods, ATPPNet generates better range images as the prediction time step increases, which can also be verified by the improvement of $4.026 \% $ over SOTA (PCPNet-semantic \cite{pcpnet}) in the mean range loss.

In \autoref{table:chamfer_loss_kitti}, we evaluate the Chamfer distance on the sampled point clouds (left column) and full-scale point clouds (right column) on the KITTI test set. As we can observe, our method is having an improvement of $10.328 \% $ over SOTA on sampled point clouds and an improvement of $8.219 \% $  over SOTA on full-scale point clouds.
% We can verify from the table that our method is having an improvement of $10.328 \% $ over SOTA on sampled point clouds and an improvement of $8.219 \% $  over SOTA on full-scale point clouds. 
% leads to a smaller Chamfer distance across all the time steps compared to the other methods for both the sampled and full-scale point clouds.
% It can be observed that the performance is similar OR This observation is consistant with our results in Table 1
% This observation is consistent with our results in $\autoref{table:range_loss_comparison}$.
It can also be observed that the margin of Chamfer distance between ATPPNet and other methods increases as the prediction time step increases (i.e., farther in future). This indicates a more stable prediction of the point clouds across all the time steps as depicted in \autoref{fig:qualitative}.
% which can also be confirmed qualitatively in \autoref{fig:qualitative}.

% \begin{table}[ht]
% \begin{center}
% \begin{tabularx}{0.9\linewidth}{|>{\centering\arraybackslash}X |>{\centering\arraybackslash}X |>{\centering\arraybackslash}X |} 
%  \hline
%  Method & Mean Chamfer distance & Mean Range loss\\ [0.5ex] 
%  \hline\hline
%  %Ours  (with finetuning) & \textbf{0.6876} & 0.7252 \\ 
%  %\hline
%  TCNet \cite{mersch2022self} & 1.389 & 0.719 \\
%  \hline
%  PCPNet-semantic \cite{pcpnet} & 1.360 & 0.704\\
%  \hline
%  Ours & \textbf{0.932} & \textbf{0.598} \\ 
%  \hline
% \end{tabularx}
% \caption{Mean Range Loss and Chamfer distance on the NUScenes test set.}
% \label{table:nu_scenes_loss_comparison}
% \end{center}
% \end{table}
In \autoref{table:nu_scenes_loss_comparison}, we report the quantitative analysis of our model trained on the nuScenes dataset.
% We have also trained our model on the nuScenes training data, and show the quantitative evaluation on the nuScenes test set in \autoref{table:nu_scenes_loss_comparison}. 
ATPPNet is improving $15.05 \% $ on the mean range loss and $31.47 \% $ on the mean Chamfer distance over SOTA.
%
% Our inference time on the KITTI dataset is $89.5$ ms and on the nuScenes dataset is $70.7$ ms.
Our inference time on the KITTI and nuScenes dataset is $89.5$ and $70.7$ ms respectively.

\subsection{Ablation Study}

% In this section, we have done a thorough ablation study on the effects of different blocks of our architecture on the KITTI test-set.
% The convolutional encoder block generates $L$ feature tensors for each of the $M$ range images. The first $L-1$ of these feature-tensors are modelled using Conv-LSTMs. To speed-up our network, the $L$-th feature tensor is modelled using $3D$-CNN. 
In this section, we conduct a thorough investigation of the relevance of different blocks of our architecture on the KITTI test set and demonstrate the effectiveness of our method.

\textbf{Impact of Attention Module}: 
In \autoref{table:abblation} column $A$, we show the results of our ablation study on the attention module.
% on the KITTI odometry test sequence $08-10$.
% In $\autoref{table:abblation}$ column "No attention", we show the performance of our architecture without  using any attention. 
To conduct this study, we set up $3$ experiments: (1) removing the attention module (column ``No Attention"), (2) using just spatial attention (column ``S-Attention") and (3) using just channel-wise attention (column ``C-Attention").
% \begin{enumerate}
%     \item removing the attention module (column ``No Attention") \label{attention_1}
%     \item using just spatial attention (column ``S-Attention") \label{attention_2}
%     \item using just channel-wise attention (column ``C-Attention"). \label{attention_3}
% \end{enumerate}
%
We observe in all the 3 experiments that the range loss and Chamfer distance deteriorates as compared to our original method.
% In experiment \ref{1} (1), we remove the attention module (column ``No Attention"), in experiment (2), we use just spatial attention (column ``S-Attention") and in experiment (3), we use just channel-wise attention (column ``C-Attention"). 
% % We observe that in all 3 experiments, the range loss and chamfer distance deteriorates as compared to our oroginal method.
% We observe that in experiment (1), while the $\mathcal{L}_C$ deteriorates slightly, the $\mathcal{L}_R$ deteriorates significantly. From experiments (2) and (3), we conclude that while the performance deteriorates compared to jointly using spatial and channel-wise attention, the deterioration is almost similar for both S-attention and  C-attention.

% Next we conduct a study on the attention module. We first remove the attention module (column "No Attention" in \autoref{table:abblation}) and observe that while the $\mathcal{L}_R$ deteriorates slightly, the $\mathcal{L}_C$ deteriorates significantly. Since we use spatial and channel-wise attention to refine the feature tensors, we conduct experiments by using just spatial attention (column "S-Attention" in \autoref{table:abblation}) and just channel-wise attention (column "C-Attention" in \autoref{table:abblation}). The observations conclude that while the performance deteriorates compared to using both spatial and channel-wise attention together, the deterioration of the performance is almost similar for both S-attention and  C-attention.

% For the first ablation study, we 

\textbf{Effects of Spatio-Temporal Modelling}:
In \autoref{table:abblation} column $B$, we demonstrate the impact of varying the number of feature tensors from the convolutional encoder to be modelled spatio-temporally. For this, we adopt three experimental setups: (1) modelling only the $L-1$-th feature tensor with Conv-LSTM and $L$-th tensor with $3D$ CNN. (column ``$L-1$ CLSTM"), (2) modelling only the $L-1$-th and $L-2$-th feature tensor with Conv-LSTM and $L$-th tensor with $3D$ CNN. (``$L-1$ \& $L-2$ CLSTM"), and (3) modelling all the $L$ layers from the convolutional encoder with Conv-LSTM (column ``all $L$ CLSTM").
% \begin{enumerate}
%     \item modelling only the $L-1$-th feature tensor with Conv-LSTM and $L$-th tensor with $3D$ CNN. (column ``$L-1$ CLSTM")
%     \item modelling only the $L-1$-th and $L-2$-th feature tensor with Conv-LSTM and $L$-th tensor with $3D$ CNN. (``$L-1$ \& $L-2$ CLSTM")
%     \item modelling all the $L$ layers from the convolutional encoder with Conv-LSTM. (column ``all $L$ CLSTM") \label{l_clstm}
% \end{enumerate}
%
For the aforementioned experiments we observe a deterioration in the performance compared to our original method. We can also conclude the importance of the $3D$-CNN layer for the $L$-th layer from experiment (3). This verifies that $3D$ CNNs are better at modelling contiguous areas in the feature tensors \cite{manttari2020interpreting} which tend to appear at the lower level features from the convolutional encoder \cite{zeiler2014visualizing}, \cite{yosinski2015understanding}.

\textbf{Impact of Sequence Length}:
In \autoref{table:seq_length}, we report the results by varying the temporal window size.
% In \autoref{table:seq_length} we demonstrate the impact of the temporal window size to our network. 
It is important to note that the temporal window size is kept the same for both input and output. We can observe a decrease in the Chamfer distance as we increase the window size from $3$ to $5$. However, further increasing the window size from $5$ to $7$ leads to an increase in the Chamfer distance. 
A possible explanation for this is that the window size of $3$ is too short to model the spatio-temporal context of the scene while, for the window size of $7$, the context length and the prediction horizon is too long.

\subsection{Results on Downstream Task}\label{downstream_task}

In this section, we analyze the impact of our model on a downstream task of generating motion estimates (i.e., odometry) for the ego vehicle.
% In this section, we analyze the quality of our predicted point clouds for generating motion estimates for the ego vehicle. 
% One approach to achieve this task is to find good feature correspondences between consecutive point clouds in real-time. 
We adopt LOAM\cite{behley2019iccv}, 
% an approach 
% for real time motion estimation and mapping 
and evaluate the disparity between the motion estimates on ground truth and predictions.
% LOAM\cite{behley2019iccv} is an approach for real time motion estimation and mapping using point clouds, 
% obtained from a 3D lidar
% which works on matching features between consecutive point clouds.
%where feature points located on planar surfaces and edges are matched to line segments and surface patches%. 
% In order to get motion estimates with minimal drift, point clouds should have quality features.
% For real time LiDAR based Odometry and Mapping, point clouds should have quality features to get an accurate transformation estimation. In this section,
 % In this section, we analyze the performance of our predicted point clouds using LOAM to generate odometry poses of our ego-vehicle. 
% In this section, we analyze the performance of our predicted point clouds in generating motion estimates for the ego using LOAM to generate odometry poses of our ego-vehicle.
% In this section, we analyze the accuracy of our predicted point clouds in generating motion estimates for the ego vehicle 
% using LOAM.
Let $\hat{p}^{\tau}\in \mathbb{R}^2$ denote the trajectory pose obtained using predicted point clouds and $p^{\tau} \in \mathbb{R}^2$ denote the trajectory pose obtained using ground truth point clouds at time step $\tau$, where $\tau \in [t+1, \dots, t+N]$.
The pose error $\mathcal{L}_P^{\tau}$ is given as:
$$
\mathcal{L}_P^{\tau} =  \mid \mid \hat{p}^{\tau} - p^{\tau} \mid \mid_2 ^2.
$$
As reported in \autoref{table:loam_pose_error}, the pose error ($\mathcal{L}_P^{\tau}$) for ATPPNet is the least as compared to the other methods. 
%  We use LOAM 
% to get the transformations between the predicted point clouds that leads to a trajectory of predicted poses $\hat{p}^{\tau}\in \mathbb{R}^2$ of the ego-vehicle where $\tau \in [t+1, \dots, t+N]$ . 
% % trajectory obtained using the input and target ground truth point clouds, 
% This trajectory of poses obtained using the predictions is compared against the trajectory of poses $p^{\tau} \in \mathbb{R}^2$ obtained using the ground truth point clouds. We compute the average pose error $\mathcal{L}_P^{\tau}$ as 
% $$
% \mathcal{L}_P^{\tau} =  \mid \mid \hat{p}^{\tau} - p^{\tau} \mid \mid_2 ^2
% $$
% % as input to LOAM. 
% We conclude from \autoref{table:loam_pose_error} that $\mathcal{L}_P$ for our method is least as compared to the other methods. 
% This means that the odometry poses obtained from our predicted point clouds are closest to the odometry poses from the ground-truth point clouds compared to the other methods. 
This verifies that the improved point cloud prediction from our proposed method translates to a tangible outcome in the form of improved localization vis-a-vis other methods. Additionally, such prediction of localization error can be effectively leveraged by active localization strategies \cite{omama2022drift} that steer the vehicle to regions where the localization is expected to be better.

% \Madhav{May want to add, "The odometry results signifies the following. Firstly it verifies the improved point cloud prediction due to the proposed method translates to a tangible outcome in the form of improved localization on the predicted point clouds vis a vis other methods. Second, such prediction of localization error can be effectively leveraged by active localization strategies \cite{IROS22-Omama} that steer the vehicle to regions where the localization is expected to be better"}

%%%%%%%%%%%%%%%%%%%%%%%%%%%%%%%%%%%%%%%%%%%%%%%%%%%%%%%%%%%%%%%%%%%%%%%%%%%%%%%%
\section{Conclusion}

In this paper, we present a novel self-supervised approach for predicting future point cloud sequences based on given past point cloud sequences. We leverage the spatial and channel-wise attention with Conv-LSTMs dually complemented by a 3D-CNN branch to model spatio-temporal information to successfully predict future point cloud sequences while operating at a frequency less than that of a LiDAR. We evaluate our method on different real-world datasets and conduct thorough ablations on our approach. Our experiments suggest that ATPPNet outperforms all other methods. We also present an application study of the predicted point clouds which highlights the potential of our approach. In future work, point clouds could be predicted by querying a location at a particular time, to determine areas with minimal drift in motion.

\bibliographystyle{IEEEtran}
\bibliography{references}
\end{document}